\documentclass[letterpaper, 10 pt, conference]{ieeeconf}  

\usepackage{graphicx}
\usepackage{xcolor}
\usepackage{amsmath, amssymb}
\newcommand*\diff{\mathop{}\!\mathrm{d}}
\newcommand{\arcangle}{%
	\mathord{<\mspace{-9mu}\mathrel{)}\mspace{2mu}}%
}

\IEEEoverridecommandlockouts                              

\title{\LARGE \bf
	UAV-Supported Maritime Search System: Experience from Valun Bay Field Trials}

\author{Stefan Ivić$^{1*}$, Luka Lanča$^{1}$, Karlo Jakac$^{1}$, Ante Sikirica$^{2}$, Stella Dumenčić$^{1}$,\\ Matej Mališa$^{1}$, Zvonimir Mrle$^{1}$, and Bojan Crnković$^{3}$ 
	\thanks{$^{1}$Faculty of Engineering, University of Rijeka, Vukovarska 58, 51000 Rijeka, Croatia}
	\thanks{$^{2}$Center for Advanced Computing and Modeling, University of Rijeka, Trg braće Mažuranića 10, 51000, Rijeka, Croatia}
	\thanks{$^{3}$Faculty of Mathematics, University of Rijeka, Ul. Radmile Matejčić 2, 51000, Rijeka, Croatia}
	\thanks{$^{*}$Corresponding author (stefan.ivic@uniri.hr)}
}

\begin{document}

\maketitle
\thispagestyle{empty}
\pagestyle{empty}

\begin{abstract}
This paper presents the integration of flow field reconstruction, dynamic probabilistic modeling,  search control, and machine vision detection in a system for autonomous maritime search operations. Field experiments conducted in Valun Bay (Cres Island, Croatia) involved real-time drifter data acquisition, surrogate flow model fitting based on computational fluid dynamics and numerical optimization, advanced multi-UAV search control and vision sensing, as well as deep learning–based object detection. The results demonstrate that a tightly coupled approach enables reliable detection of floating targets under realistic uncertainties and complex environmental conditions, providing concrete insights for future autonomous maritime search and rescue applications.
\end{abstract}

\section{Introduction}
Searching for and localizing targets at the sea surface, especially in the context of search and rescue or maritime hazard identification, remains a challenging task, often hampered by environmental unpredictability and technological constraints. Traditional methods, relying on extensive manual effort, are limited by slow coverage and subject to observational bias. Recent advances in sensor integration, autonomous vehicles, and data-driven flow modeling offer new opportunities to design efficient, adaptive, and robust search strategies.

This paper reports on a comprehensive experimental field campaign aimed at testing a novel methodology for autonomous aerial search. The approach synergistically combines real-time measurement and reconstruction of sea surface flow fields, adaptive probabilistic search planning, and deep-learning-based vision for onboard detection within unmanned aerial vehicles (UAVs). The main contributions include an in-depth analysis of the coupled system's performance under the uncertainties of a real marine operation, evaluation of the field estimation, search and detection subsystems, and detailed insights into integration strategies that maximize the real-world effectiveness of such automated search missions.

\section{Literature Review}
Maritime search and rescue (SAR) operations are highly challenging due to the vast areas that need to be covered and the constantly changing ocean conditions. Unmanned aerial vehicles (UAVs) have proven useful as they enable rapid area exploration, reduce risks for human crews, and can be equipped with visual and thermal sensors for target detection. However, their application is not straightforward, as limitations in communication range and flight coordination remain critical issues. An additional complexity arises from the motion of search fields caused by ocean currents, which increases uncertainty and complicates mission planning. Consequently, there is a strong need for methods that account for the dynamic nature of the maritime environment and allow real-time adaptation of search strategies, thereby improving efficiency and increasing the probability of successful target detection and localization.

Reconstruction of near-surface ocean velocity fields represents a crucial step for understanding and predicting the dynamics of drifting targets in maritime environments. Various approaches have been developed to estimate these fields by combining models and observations. Data assimilation techniques integrate satellite and in situ measurements into consistent ocean state estimates, enabling both real-time forecasting and retrospective reanalysis, with recent advances incorporating machine learning to improve efficiency and accuracy \cite{martin2025data}. Complementary to model-based approaches, large-scale velocity fields can be reconstructed directly from drifter observations. Drifters are an indispensable tool for oceanographic research, as they enable direct observations of dispersion processes in the upper ocean, which cannot be adequately measured by other instruments. Their applications extend to climate studies, oil spill tracking, iceberg monitoring, and SAR operations \cite{lumpkin2013global,lumpkin2017advances}. The drifter measurements provide the basis for reliable real-time reconstruction of the surface flow that is essential for planning and accomplishing effective SAR missions at sea.

Conducting search missions in dynamic maritime environments requires UAV motion control adjusted for sea dynamics, environment perception and computer vision target sensing and detection. A key approach involves the use of probability maps that evolve over time to reflect both environmental dynamics and sensing outcomes. For instance, solar-powered UAVs have been employed with Gaussian mixture models to estimate prior target distributions, enabling dynamic task allocation and real-time probability map updates that account for environmental velocity fields \cite{lun2022target}. At the decision-making level, optimization-based algorithms have been proposed to enhance SAR planning, focusing on resource scheduling, task allocation, and route optimization. Genetic simulated annealing and spatio-temporal task allocation methods have shown promise in producing more efficient emergency response plans compared to traditional experience-based approaches \cite{ai2019intelligent}. Beyond motion planning, advances in automatic detection systems have significantly contributed to maritime SAR. UAVs equipped with optical sensors, combined with deep learning techniques, enable the identification of individuals at sea from aerial imagery, with synthetic data increasingly used to train robust detection models \cite{martinez2025maritime}. Finally, recent work has focused on ergodic search in dynamic environments, where evolving probability fields are coupled with potential-based multi-agent control. This framework allows multiple agents to adapt their search trajectories to dynamic target distribution influenced by the flow field, providing improved efficiency and accurate in simulated synthetic and realistic SAR scenarios \cite{lanvca2025ergodic}. The reviewed approaches highlight the importance of integrating motion control, intelligent planning, target dynamics, and advanced detection systems for successful search operations in complex maritime conditions.

\section{Methodology}
This section outlines the comprehensive methodological framework developed to enable the maritime UAV search experiment. The UAV motion control relies on real-time dynamics of target probability distribution governed by the surface flow reconstructed from drifter measurements. The probabilistic search strategy also integrates the UAV sensor characteristics and machine-learning detection system performance.

\subsection{Surface Flow Reconstruction}

Developing a realistic surface flow representation is critical for predicting the drift and potential locations of floating targets. A variety of techniques exist for monitoring or modeling sea surface circulation, each with its own advantages and drawbacks. In real-world maritime search operations that demand rapid deployment and flexible coverage, drifters provide a useful compromise by combining real-time measurements, ease of deployment, and reliability, although they are limited to providing sparse point-based observations. This limitation can be addressed with advanced flow reconstruction techniques that often require extensive computational resources and time. 

Due to time limitations of SAR operations, a simplified surrogate model fitting is chosen for flow reconstruction. This framework omits additional complexities such as wind, waves, tides, and thermal variations, yet preserves the essential flow dynamics through computational fluid dynamics (CFD). The approach adopted from \cite{JAKAC2025104699} combines two simplified two-dimensional flow models: one using a domain with coastal and inlet/outlet boundaries, and the second using a circular domain with the inlet/outlet condition throughout the entire boundary. These are solved independently and then merged into a hybrid field capable of reproducing submesoscale circulation and passive tracer transport, such as drifting objects or pollutants. The surrogate formulation adopts a steady, incompressible flow description over a two-dimensional region $\Omega \subset \mathbb{R}^2$, governed by the steady incompressible Navier--Stokes equations \cite{gunzburger2012finite,lions1996mathematical}:
\begin{equation*}
	\rho (\mathbf{w} \cdot \nabla) \mathbf{w} = -\nabla p + \mu \nabla^2 \mathbf{w} + \rho \mathbf{f}
	\label{eq:ns_equation}
\end{equation*}
\begin{equation*}
	\nabla \cdot \mathbf{w} = 0.
	\label{eq:compressibility_equation}
\end{equation*}
Here, $\mathbf{w}$ denotes the velocity field, $p$ the dynamic pressure, $\rho$ the fluid density, $\mu$ the dynamic viscosity, and $\mathbf{f}$ the external body forces. The incompressibility condition assumes that the density $\rho$ remains constant.

The flow is modeled using a pressure-driven approach, which is particularly effective when the precise inlet and outlet flow realization is unknown. In other words, inlet and outlet regions of the domain boundary are not set in advance but are determined as part of the resulting flow. In the flow model computing procedure, the boundary conditions are iteratively adjusted to zero-gradient velocity for outlet, and to the flux normal components for inlet boundaries. This requires specifying total pressure $p_0$ along the boundary, along with the additional tangential velocity component $\mathbf{w}_t$. The dynamic pressure is then evaluated depending on whether the boundary acts as an inlet or an outlet:
\begin{equation*}
	p = 
	\begin{cases}
		p_0 & \text{for an outlet,} \\
		p_0 - 0.5 \rho |\mathbf{w}|^2 & \text{for an inlet.}
	\end{cases}
	\label{eq:pressure_calculation}
\end{equation*}

To estimate a complete and realistic velocity field $w$ over the search domain, two separate OpenFOAM-based simulations are combined. A 'bounded flow' representing a coastal region with features such as coastlines and inlet/outlet boundaries, and a  'open flow' on a circular wall-less domain to capture broader environmental influences.

By combining the resulting bounded and open flow velocity fields, the model offers a more complete and flexible approximation of the surface flow behavior. In real marine environments, atypical flow behavior, such as currents that appear to originate from land, has been observed in studies \cite{bellomo2015toward, berta2014surface}. Standard two-dimensional CFD approaches cannot accurately capture this flow behavior because they cannot account for complex three-dimensional coastal variability. However, the fused model is designed to allow for both typical and atypical flow patterns which can be fitted to real-time measurements, providing a surrogate model feasibility to match the actual coastal and marine conditions.

The surface velocity field reconstruction is addressed as a model fitting problem with the primary objective of aligning a simulated flow field with measurements obtained at scattered points. The goal is to determine control values for the boundary conditions interpolation that simulates flow closely matching the real-world observations. The fitting is performed using an optimization algorithm, which iteratively updates the boundary condition control values to minimize the difference between the simulated and measured velocities at the sampling locations.

In numerical implementation, the boundary conditions are smoothly defined along the inlet/outlet edges by using the interpolation on a set of control points. Each control point specifies two quantities: tangential velocity ($w_t$) and pressure ($p$). Cubic spline interpolation is then applied to generate pressure and velocity profiles at the boundary cells. This maintains continuity while reducing the number of variables required to define the boundary conditions.

The optimization vector $\mathbf{b}$ contains the tangential velocity and pressure values at each boundary control point for both base models:
\begin{equation*}
\mathbf{b} = \left(\mathbf{w}_{t,1}, p_1, \ldots, \mathbf{w}_{t,n_{CP}}, p_{n_{CP}}\right)^T,
\end{equation*}
where $n_{CP}$ denotes the total number of boundary control points. It defines any trial flow in the observed domain.

Each evaluation of a trial flow defined with $\mathbb{b}$ includes two OpenFOAM simulations, each for 'bounded' and 'open' flow model, resulting with velocity vector fields $\mathbf{w}_b$ and $\mathbf{w}_o$, respectively. They are aggregated to provide a fused velocity field used as a surrogate flow model
\begin{equation*}
	\mathbf{w}_s = \mathbf{w}_b + \mathbf{w}_o.
\end{equation*}

The error of the fitted model $E_d$ is assessed as the root mean square of the velocity differences magnitude at the sampling points. The optimization goal is to minimize $E_d$, ensuring the simulated stationary velocities align as closely as possible with the reference data. This procedure is repeated over the interval $T_u$, which represents the expected period between successive measurement updates. In this way, unsteady flow can be approximated as a sequence of steady-state solutions that adapt to transient alterations.
This quasi-steady assumption relies on the fact that many oceanic flows evolve slowly and gradually, making them approximately steady over short time intervals. As a result, the flow model used in the optimization process remains synchronized with the actual environmental dynamics.

\begin{figure}[thpb]
	\centering
	\includegraphics[width=1.0\linewidth]{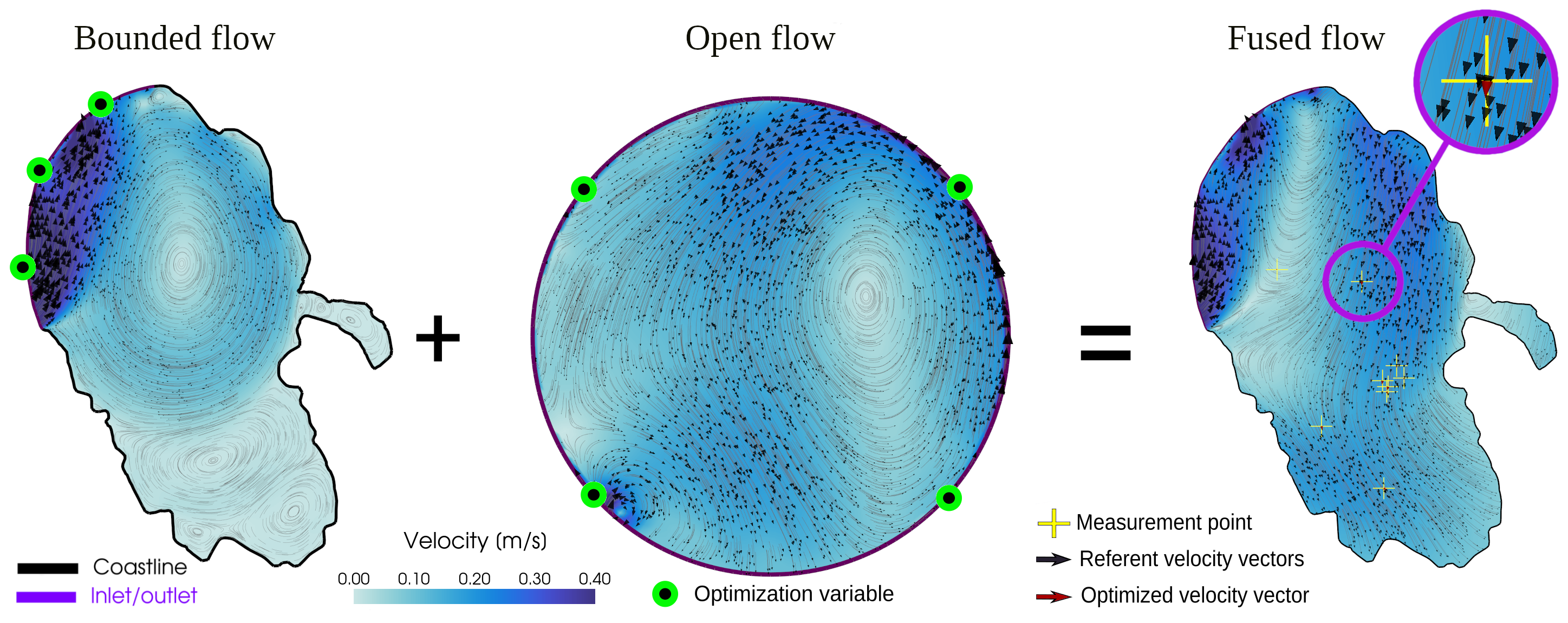}
	\caption{Illustration of the fused flow surrogate model concept combining bounded and open velocity fields. The optimization vector consists of $p$ and $\mathbf{w}_t$ values at boundary condition control points. The figure also displays measured and fitted velocity vectors at the drifter locations, used to establish the model fitting error.}
	\label{fig:fused-flow-fields}
\end{figure}

The presented surrogate model fitting problem can be formulated as an optimization problem
\begin{equation*}
	\begin{aligned}
		& \underset{\mathbf{b}}{\text{minimize}}
		& & E_d(\mathbf{b}) =  \sqrt{\dfrac{1}{n_{D}} \sum_{i=1}^{n_{D}} \left|\left|\mathbf{w}_{r,i} - \mathbf{w}_{s,i}(\mathbf{b})\right|\right|^2}\\
		& \text{subject to}
		& & \mathbf{b}_l \leq \mathbf{b} \leq \mathbf{b}_u.
	\end{aligned}
	\label{eq:drifter_error}
\end{equation*}
The index and number of drifter measurement locations are denoted with $i$ and $n_{D}$, respectively, while $\mathbf{w}_{r}$ represents the measured reference velocity, and $\mathbf{w}_{s}$ is the simulated velocity at the corresponding sampling location. An illustration of the proposed approach using surrogate fused flow model fitting is presented in Figure~\ref{fig:fused-flow-fields}.

To solve the optimization problem, the widely used Particle Swarm Optimization (PSO) algorithm is employed to adjust the optimization vector, enabling better capture of complex flow features while reducing the risk of convergence to local minima. It should be noted that the flow field reconstruction is not limited by a fixed number of optimization iterations. Instead, it is performed over a predefined time interval $T_u =$ 10 minutes. This interval is chosen to ensure that it is short enough to resolve significant flow changes, yet long enough to maintain computational efficiency for real-time application. This approach allows the system to periodically update the velocity field, assimilate new measurements as they become available, and it justifies the steady flow approximation within each reasonably short interval $T_u$.

\subsection{Drifting error assessment}

In a real marine environment, while providing velocity measurements drifters are continuously advected by the sea currents. This is often a disadvantage, since they can drift outside the region of interest or aggregate, making the measurements redundant. However, the inevitable motion can provide an advantage, namely the estimation of the drift error. After each quasi-steady update period $T_u$, a drifter trajectory is determined by integrating a previously reconstructed velocity field:
\begin{equation*}
	\frac{\diff \mathbf{z}}{\diff \tau} = \mathbf{w}_f(\mathbf{z}, t - T_u).
	\label{eq:location_advection}
\end{equation*}
Solving the motion law for $\tau \in [t - T_u, t]$ provides simulated drifter trajectory $\mathbf{z}(\tau)$ and final expected position of the drifter at time $t$. This approach ensures that simulated drifter trajectories evolve consistently with the reconstructed flow field. Over successive intervals, it produces an adaptive approximation of drifter motion, supporting timely and field-ready decision-making.

Advecting drifters also provides a way to assess the accuracy of the reconstructed flow after each period $T_u$, by comparing real measured locations and simulated on fitted flow field. This is done by computing an average positional error between the reference and simulated drifter locations, defined as
\begin{equation}
	S(t) = \frac{1}{N} \sum_{i=1}^{N} 
	\left\lVert \mathbf{z}_{\text{r,}i}(t) - \mathbf{z}_{\text{s,}i}(t) \right\rVert ,
	\label{eq:location_error}
\end{equation}
where $\mathbf{z}_r$ is the referent (measured) and $\mathbf{z}_s$ is simulated position of a drifter.

\subsection{Target Detection}
Visual detection of floating targets is performed by integrating a deep learning object detection model into the UAV's operational loop. Specifically, a YOLOv8-based artificial network, fine-tuned on an aerial maritime dataset, is used to identify and localize targets from onboard images. The design of floating targets is described in section \ref{sec:targets_and_drifters}. The dataset, gathered specifically for the experiment, includes 522 high-resolution aerial images (captured at heights of 60–100 m) and annotated across three classes: sea targets, drifters, and boats. For training and evaluation, the data is split into training, validation, and test sets in an 80:10:10 ratio to ensure generalization and prevent overfitting.

\begin{figure}[thpb]
	\centering
	\includegraphics[width=0.49\linewidth]{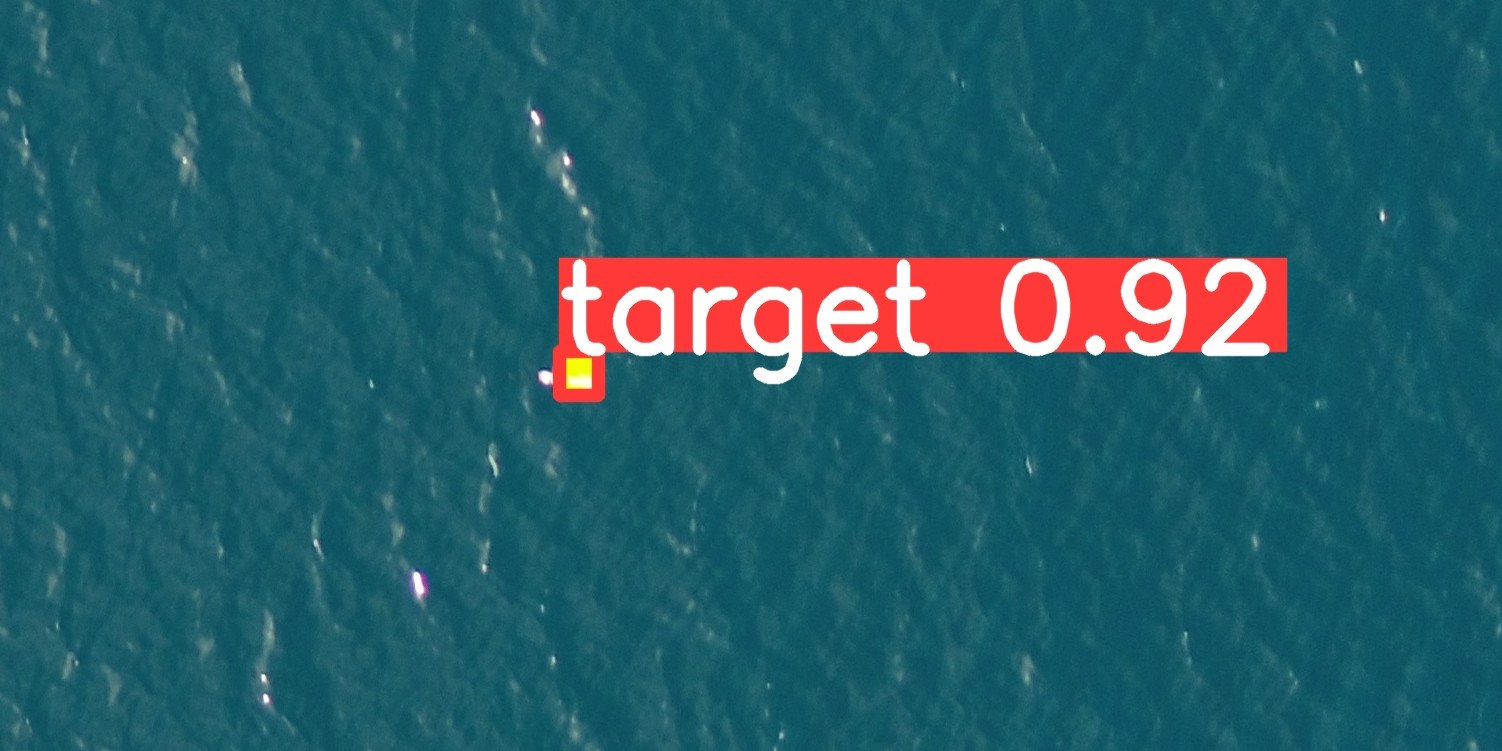}%
	\hfill
	\includegraphics[width=0.49\linewidth]{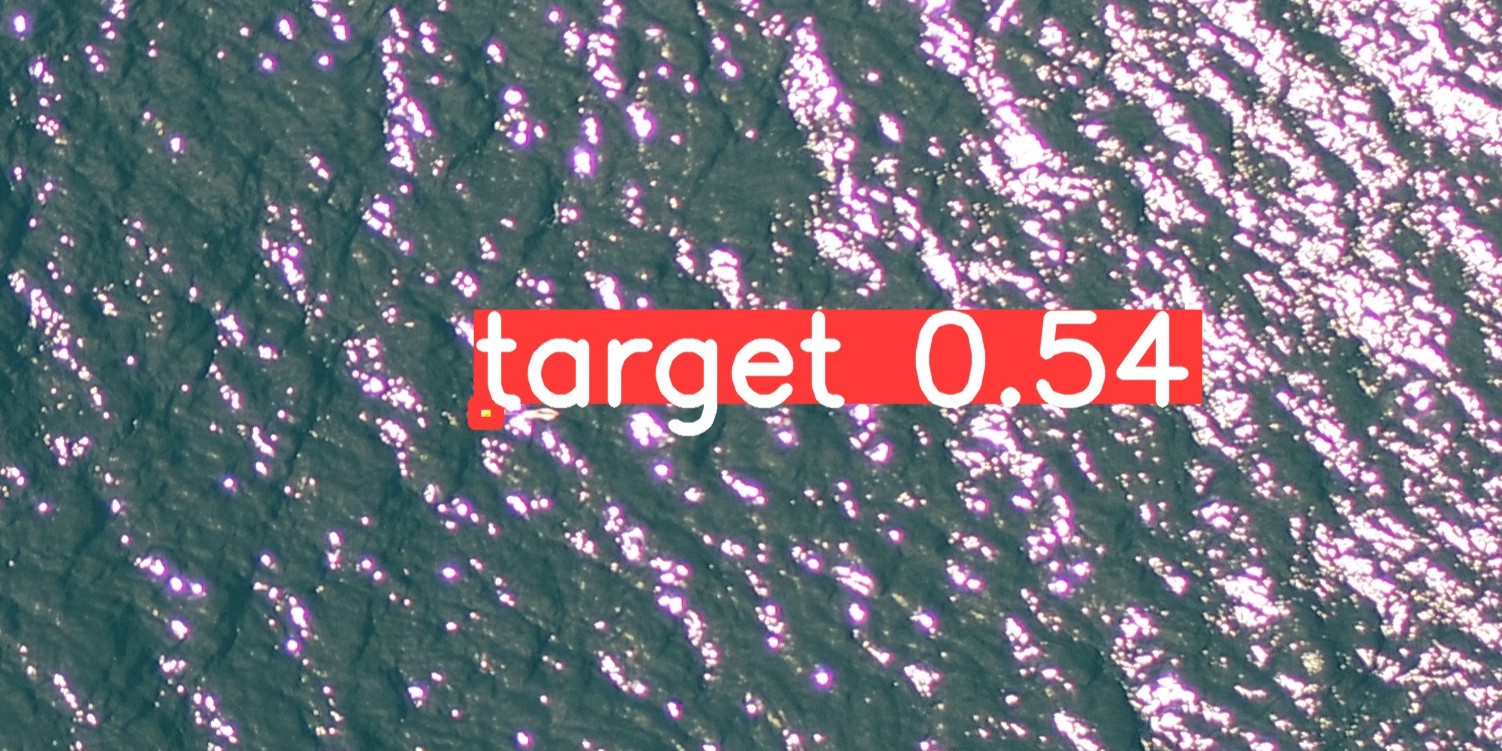}
	\\
	\vspace{2mm}
	\includegraphics[width=0.49\linewidth]{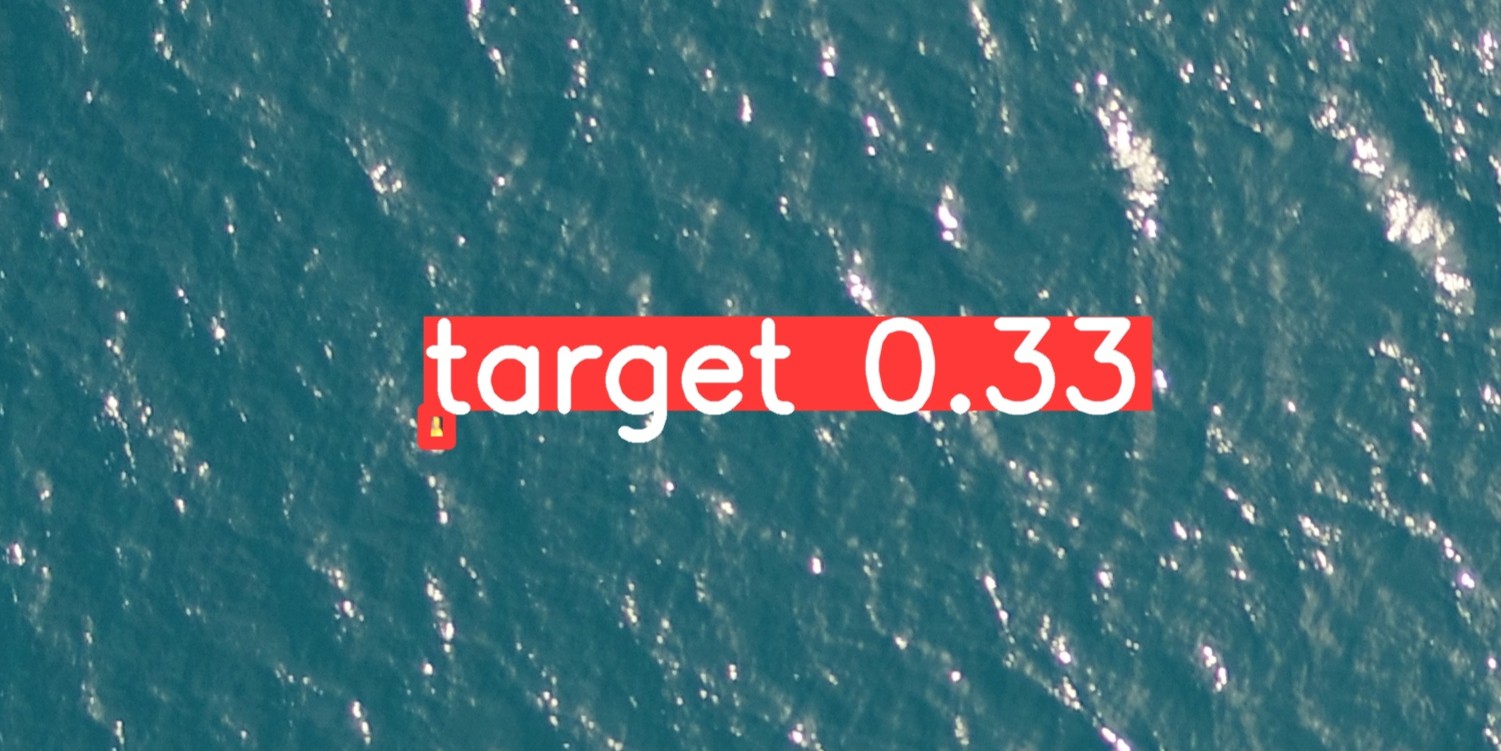}%
	\hfill
	\includegraphics[width=0.49\linewidth]{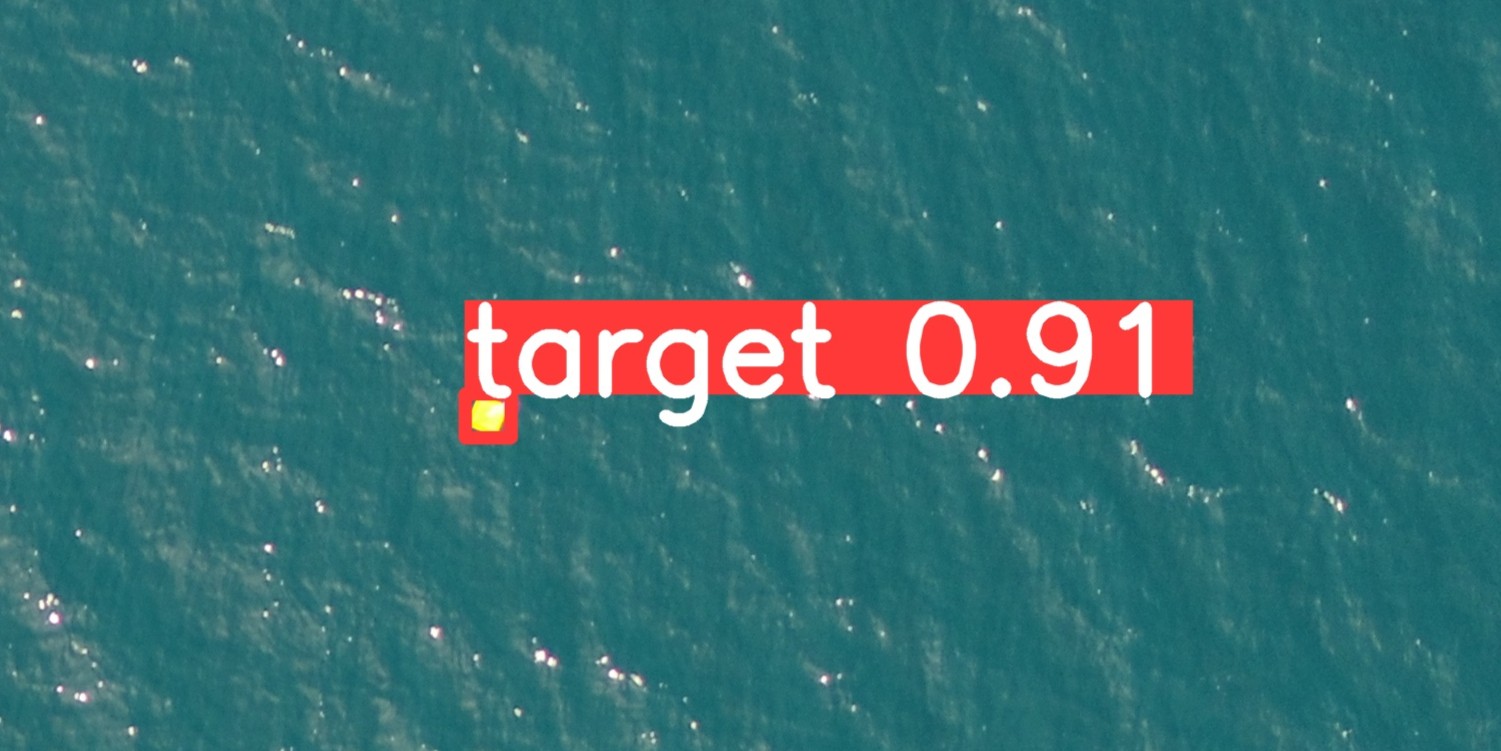}
	\caption{Example detections of targets, with associated detection confidence using YOLOv8}
	\label{fig:detections-of-targets}
\end{figure}

Focusing on the target class, which is of primary operational interest, the network was trained for 100 epochs using a pre-trained “yolo8l.pt” model as the starting point, with a batch size of 4 and images resized to 640 px for consistency. On test data, the model reached a mean average precision (mAP) of 0.723 at IoU=0.5, a precision of 0.861, and recall $r= 0.68$ at the default confidence threshold. The recall value is particularly salient for the probabilistic search framework, as it directly informs the “sensing function” as explained in the next section. Example detections of sea targets are shown in Figure~\ref{fig:detections-of-targets}, with the corresponding detection confidence values displayed next to the class labels.

\subsection{UAV Search}

Generally, when a target is lost at sea, the report specifies a belief about the target’s estimated area and, if available, the associated probability. The search mission involves inspecting the sea surface, represented by the two-dimensional domain $\Omega$, where an arbitrary point is denoted by $\mathbf{x}$. The search is guided by a dynamic probabilistic belief about the locations of undetected targets, represented by the evolving spatial distribution of the undetected target probability density $m(\mathbf{x}, t)$. Prior to the search, the initial undetected target probability density $m(\mathbf{x}, 0)=m_0(\mathbf{x})$ satisfies
\begin{equation*}
	\int_\Omega m_0 \diff\mathbf{x}=1. 
	\label{eq:m_normalization}
\end{equation*}

As time progresses, $m$ evolves due to the effects of the sea surface layer flow field and the search efforts. The affects of the flow field are modeled with advection while uncertainties are accounted for with diffusion. As the inspection progresses, portions of the domain are examined, reducing the probability that an undetected target is located in the inspected areas. As presented in \cite{lanvca2025ergodic}, the evolution of m is described with equation 
\begin{equation*}
	\frac{\partial m}{\partial t} = D\cdot \nabla^2 m - \mathbf{w}\cdot\nabla m - \Gamma\cdot m,
	\label{eq:pde_m}
\end{equation*}
where $\Gamma$ represents the instantaneous detection probability. The term $\Gamma \cdot m$ represents continuous sensing effort on the undetected target distribution. Considering the velocities of the flow field, this can be implemented discretely at every sensing interval $\Delta t_s$, since the effects of advection and diffusion are negligible over $\Delta T_u$. At each sensing time step $j$, a UAV captures an image covering the sea surface area within its camera Field of View (FOV), denoted by $\Omega_{\text{FOV}}$. The search is conducted with $n$ UAVs, where $k$ represents the index of each UAV, and the resulting sensing update at each time step is applied using the Euler scheme, described as 
\begin{equation*}
	m_{j+1}(\mathbf{x}) =
	m_j(\mathbf{x}) \cdot \prod_{k=1}^{n} 
	\begin{cases}
		1 - r, & \mathbf{x} \in \Omega_{\text{FOV,} k} \\
		1, & \text{otherwise}
	\end{cases}
\end{equation*}

The core of search control is an ergodic control method called Heat Equation Driven Area Coverage (HEDAC) \cite{ivic2016ergodicity}.
To explore the search domain according to $m$, the UAVs are guided by a potential field $u$, obtained by solving a partial differential equation
\begin{equation}
	\alpha \nabla^2 u (\mathbf{x}, t) -u (\mathbf{x}, t) + m (\mathbf{x}, t) = 0,
	\label{eq:hedac_pde}
\end{equation} 
with the boundary condition 
\begin{equation*}
	\left.\frac{\partial u}{\partial\mathbf{n}}\right|_B = 0,
	\label{eq:hedac_bc}
\end{equation*}
where $\mathbf{n}$ denotes the outward normal vector on the boundary $B$. The gradient of the potential field is obtained by
\begin{equation*}
	\mathbf{u}(\mathbf{x}) = \frac{\nabla u(\mathbf{x})}{||\nabla u(\mathbf{x})||}
\end{equation*}
which directs the search agents towards the regions of higher undetected target probability density $m$.

During the search, the UAVs fly at a constant height and utilize a constant forward velocity $v$.
The yaw angular velocity $\omega$ is determined for each UAV according to the rate of change of the angular difference between the current heading $\theta_i$ and the direction of the potential field gradient $\mathbf{u}$ at UAVs current location:
\begin{equation*}
	\omega_{i} = \frac{\diff}{\diff t} \left(\arcangle \left(\theta_i(t), \;\mathbf{u}(\mathbf{y}_i(t)) \right)\right).
	\label{eq:dubins_angular_velocity}
\end{equation*}

\subsection{Brownian motion and adaptive diffusion}
Simplified model fitted to sparse data contains inherent sources of error from idealized assumptions and limited and inaccurate measurements. To compensate for these errors the target distribution is diffused which results with a greater search area but ensuring the targets are always contained in the distribution $m$. The coefficient is calculated by comparing the difference between predicted and observed drifter locations after each period $T_u$. The diffusion coefficient is adjusted using the mean squared displacement (MSD), denoted as $S^2$. The MSD is calculated from the drifter position error defined in equation \ref{eq:location_error} and measures the average deviation due to inaccuracies in the reconstructed flow. Adjusting the diffusion in this way increases the spread of the advected scalar, improving the likelihood that it covers the targets regardless of reconstructed flow errors. Considering a two-dimensional Brownian particle motion, the connection between MSD and diffusion coefficient is given by:
\begin{equation*}
	S^2(t) = 4 \cdot D \cdot t.
\end{equation*}
Based on this, a compensating diffusion term can be calculated from drifting error emerged in period $T_u$ as:
\begin{equation}
	D = \frac{S^2}{4 \cdot T_u}.
	\label{eq:D_c}
\end{equation}
Using this adaptive diffusion increases the likelihood that the advected scalar field covers areas where the target may be located, even in the presence of unavoidable measurement errors, flow reconstruction inaccuracies, and other uncertainties in the system.

\section{Experimental Setup}
This section describes the implementation of the search system and practical details regarding the preparation of field trials.

\subsection{Location and Organization}
\begin{figure}[thpb]
	\centering
	\includegraphics[width=0.49\linewidth]{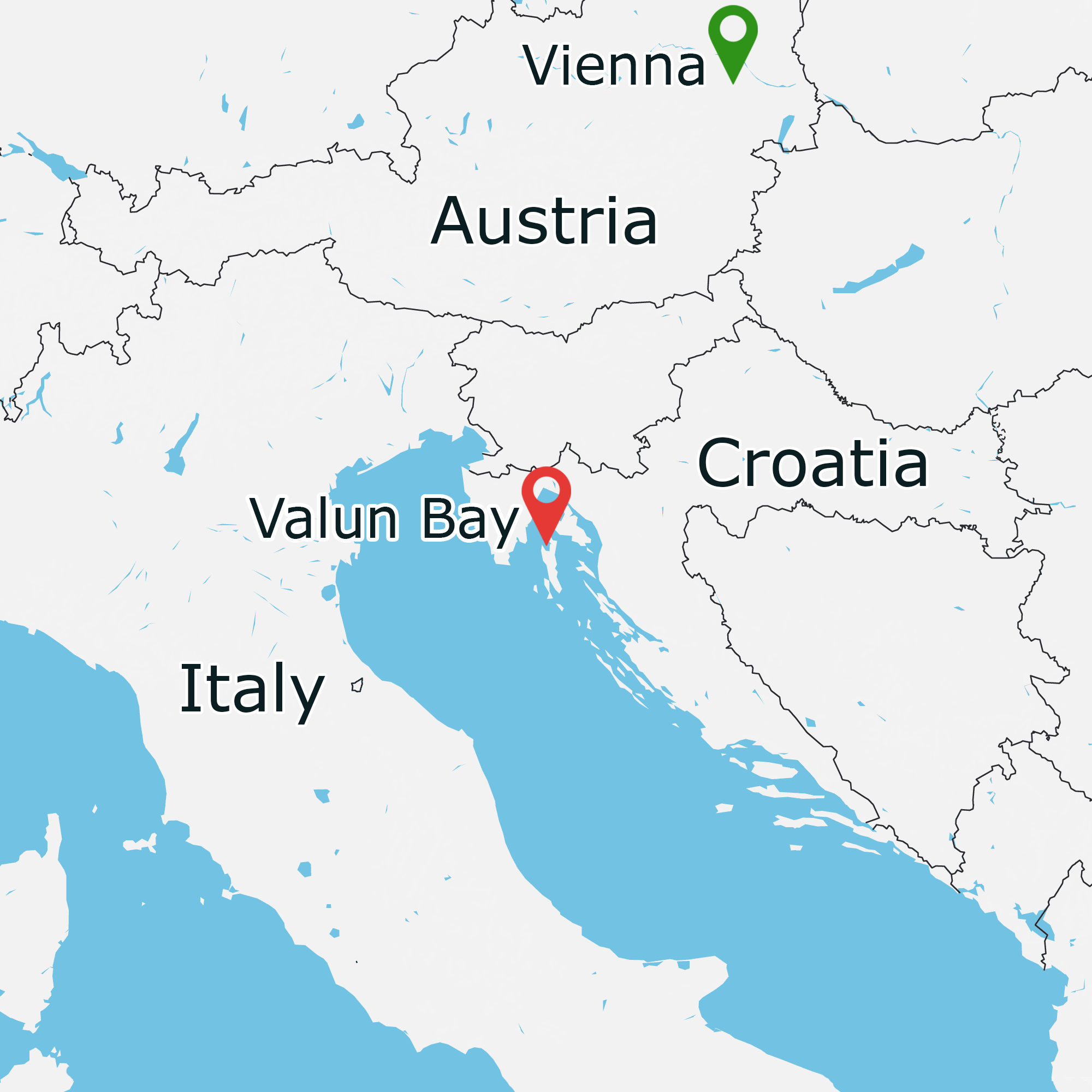}%
	\hfill
	\includegraphics[width=0.49\linewidth]{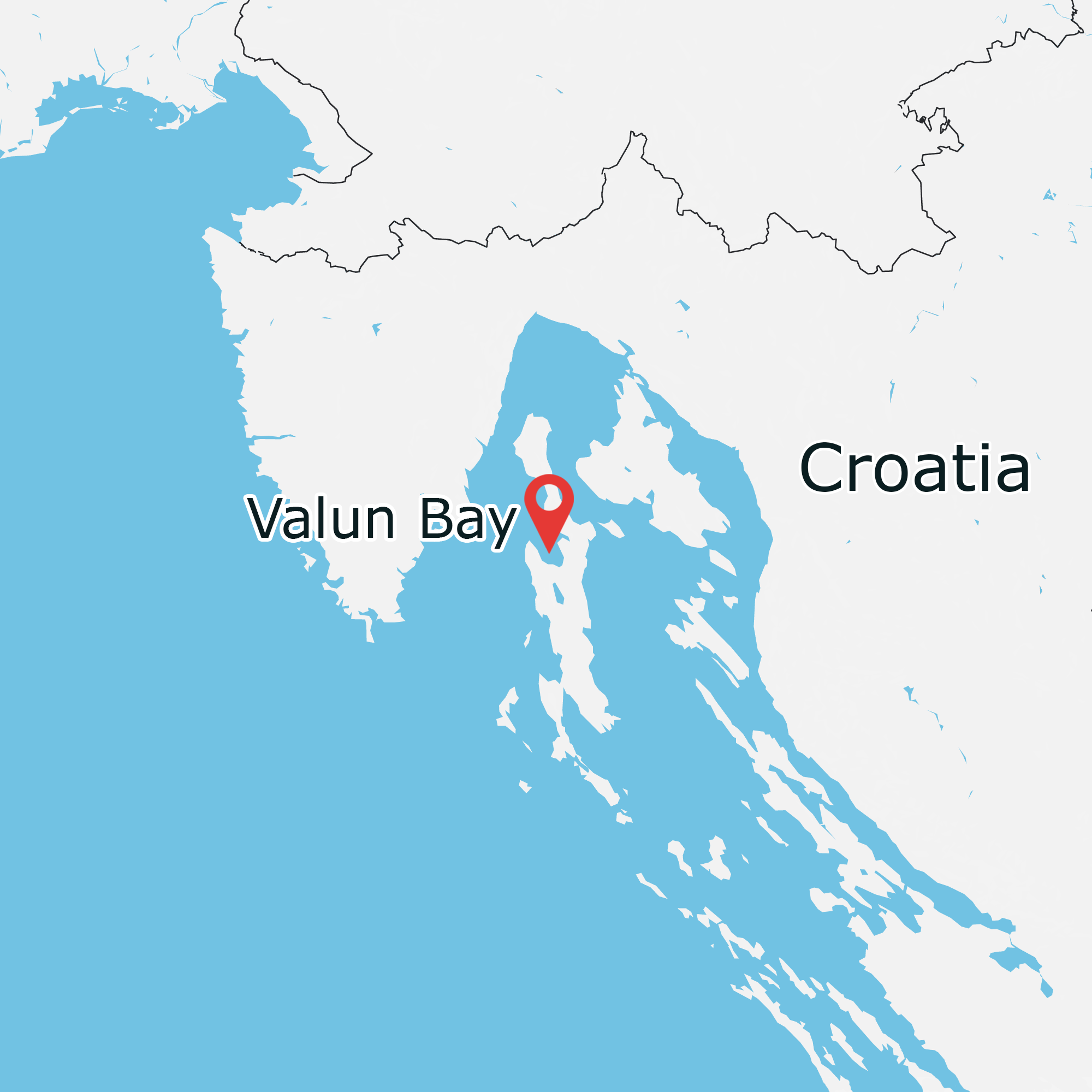}
	\caption{Map of part of Europe with marked location of the experiment (red pin) and the location of the ICRA 2026 conference (green pin).}
	\label{fig:domain}
\end{figure}

\begin{figure}[thpb]
	\centering
	\includegraphics[width=0.49\linewidth]{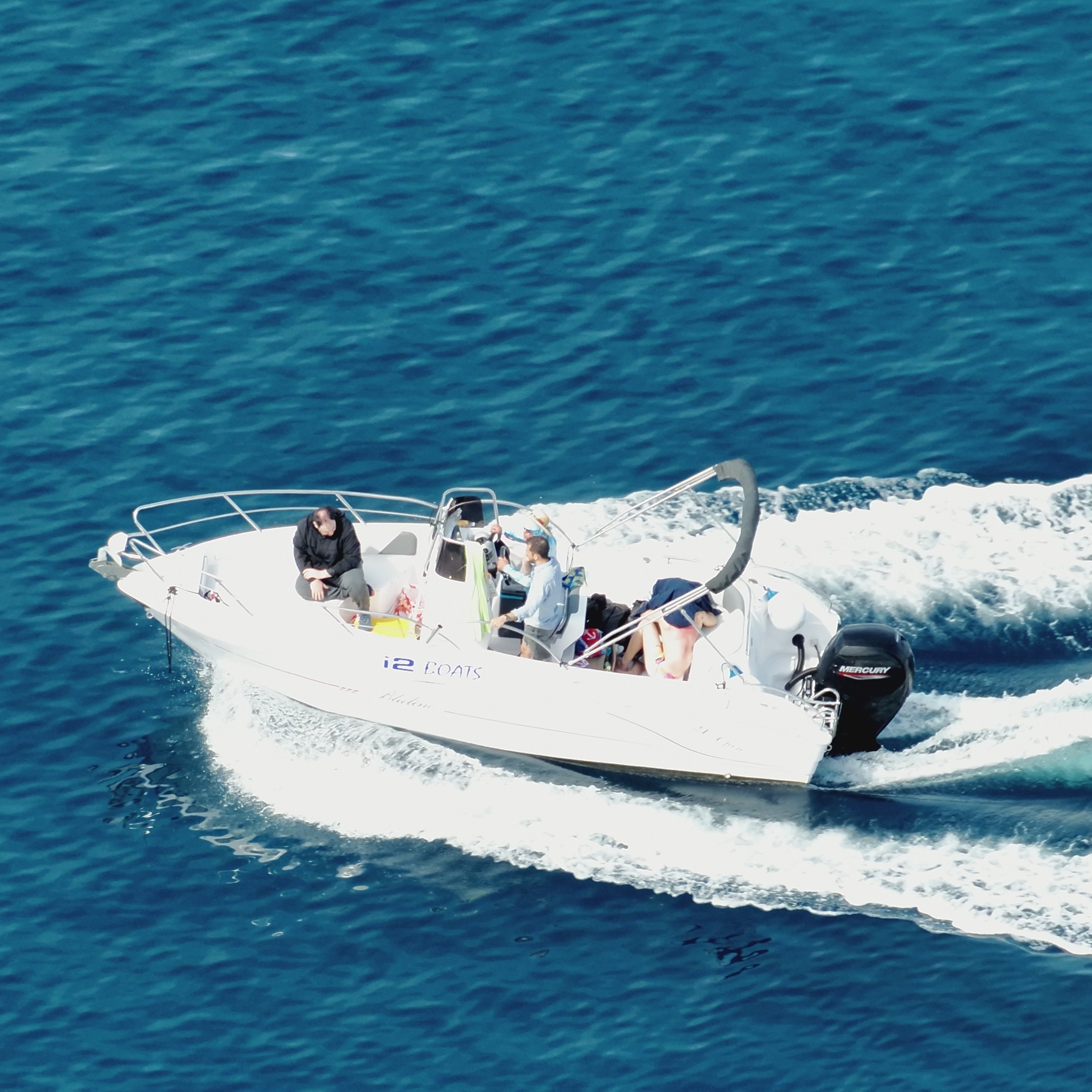}%
	\hfill
	\includegraphics[width=0.49\linewidth]{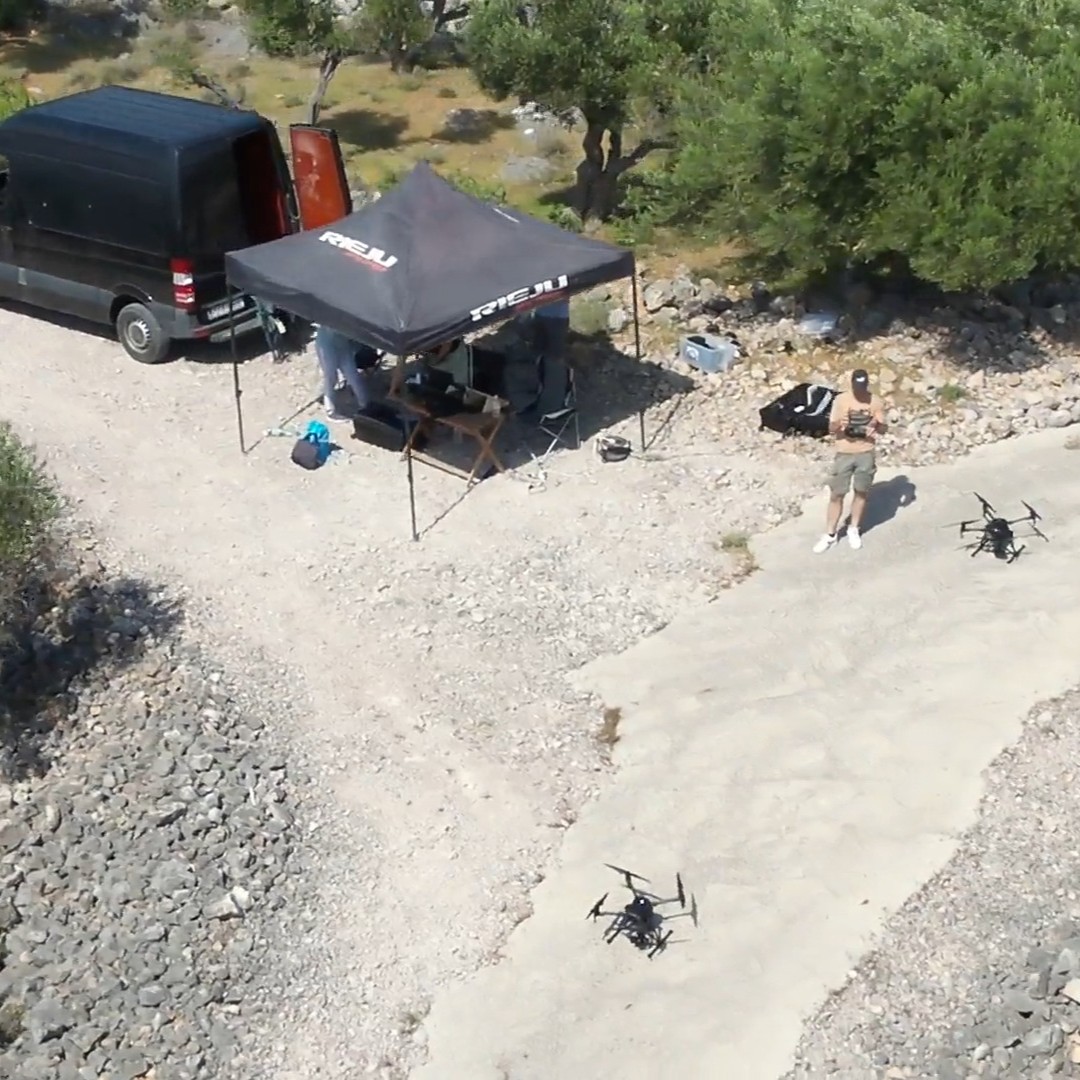}
	\caption{Sea unit deploying the experimental target (left) and UAV base station overlooking the Valun Bay search domain (right)}
	\label{fig:sea-unit-and-base-station}
\end{figure}

The field test took place in Valun Bay, on the western shores of Cres Island, Croatia, on June 4th, 2025 (see Figure~\ref{fig:domain}). The bay covers an area of approximately 55.8~km$^2$, providing a representative coastal scenario with complex shoreline geometry and variable current regimes. The operational team was divided into two main subgroups: a sea unit on the vessel (responsible for deploying targets and drifters, see Figure~\ref{fig:sea-unit-and-base-station}(left)) and a land unit stationed on the eastern bay coastline (85 m above sea level, see Figure~\ref{fig:sea-unit-and-base-station}(right)). The location of the land unit was selected to maximize both visibility and radio coverage for UAV control and uninterrupted drifter data reception.

\subsection{Targets and Drifters}
\label{sec:targets_and_drifters}

\begin{figure}[thpb]
	\centering
	\includegraphics[width=0.49\linewidth]{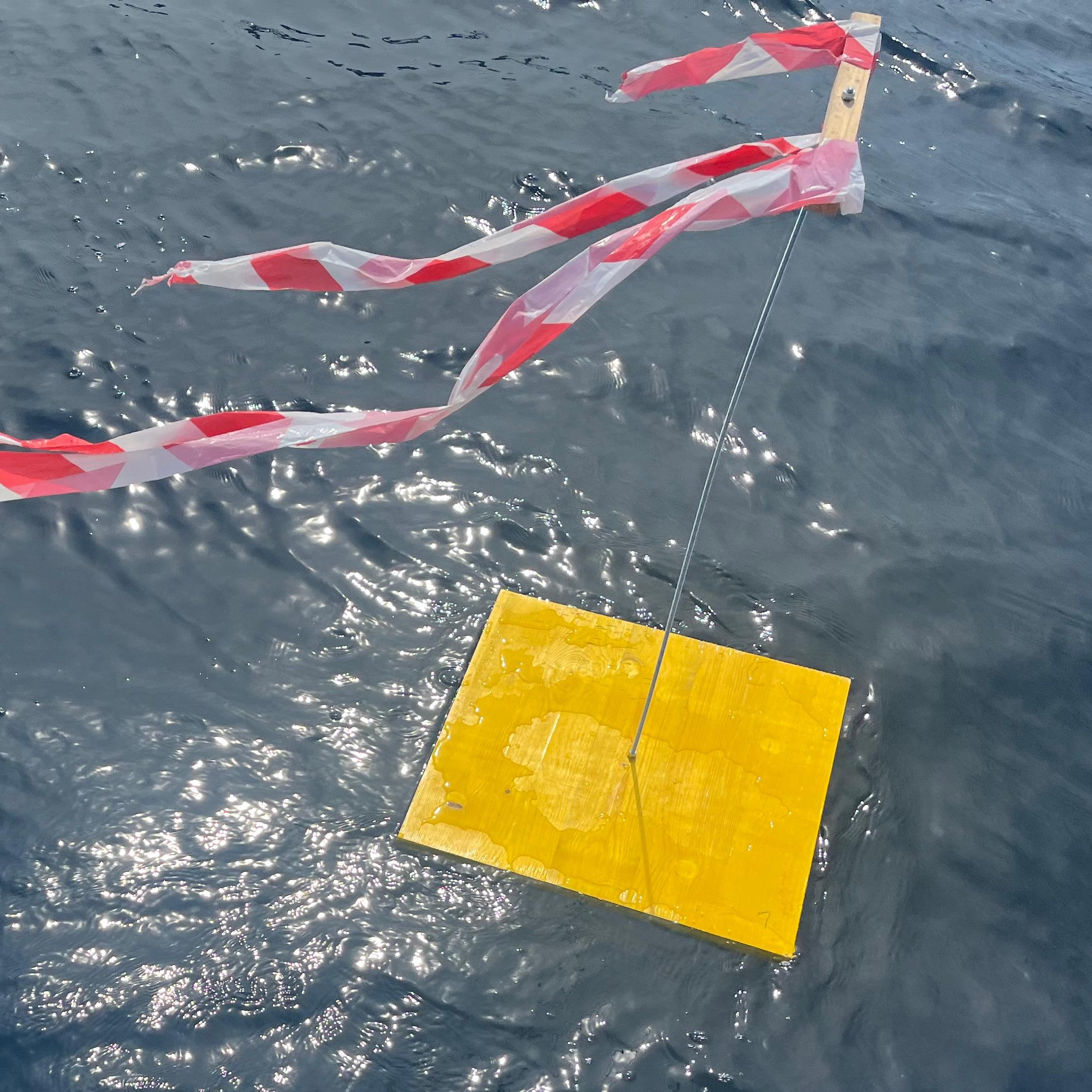}%
	\hfill
	\includegraphics[width=0.49\linewidth]{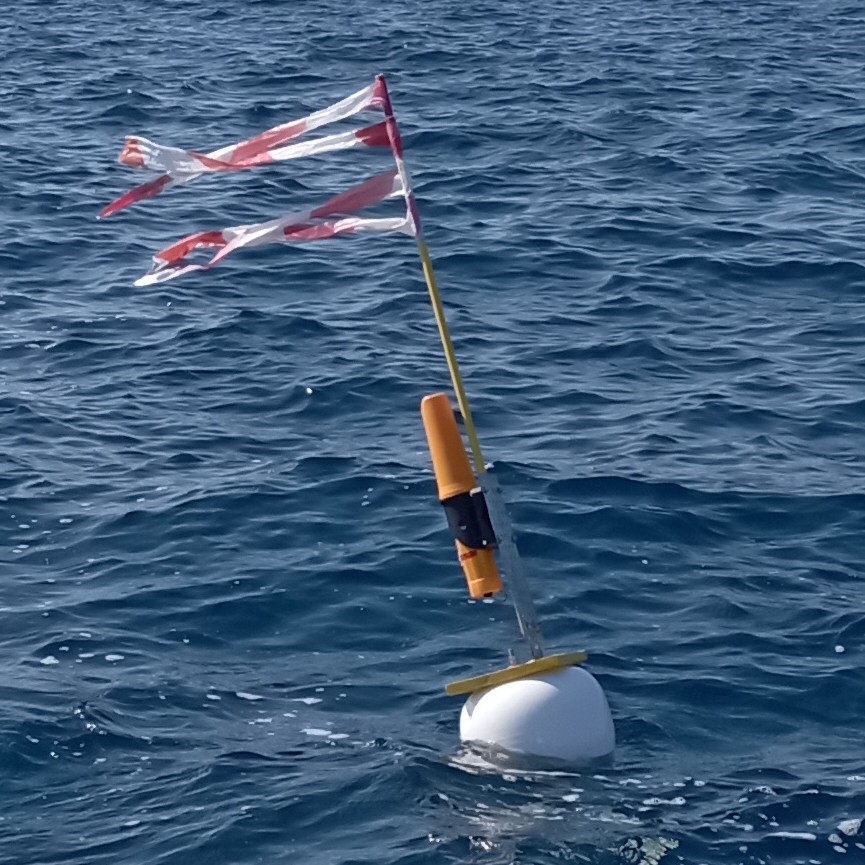}
	\caption{Example showing the floating target from the experiment (left) and the drifter used to obtain surface flow velocities (right)}
	\label{fig:target-and-drifter}
\end{figure}

A total of twelve drifters and four custom-made sea targets were used. Each target was constructed from a 0.5 $\times$ 0.5~m yellow-painted wooden board, fitted with a 1 m metal pole and attached marking tape to improve horizontal visibility from a marine unit. One example of a target is shown in Figure~\ref{fig:target-and-drifter}(left). 

The targets were deployed in two distinct regions, corresponding to two different search scenarios. The first deployment took place at 10:15, with the targets arranged in a plus-shaped pattern within a circular area of 300 m radius, centered approximately 1.4 km from the land unit. The second deployment occurred at 14:17, with the targets distributed in the same pattern within a circular area of 250 m radius. This area was located on the opposite side of the bay, approximately 3.75 km from the land unit.

Drifters were assigned three roles: four were distributed evenly throughout the bay to improve global flow field accuracy, five were deployed closer to the target region for enhanced local estimation, and three served as validation units excluded from the main optimization loop, providing ground-truth error measurement and system calibration. 

In this experiment, custom drifters -- comprising floating buoys equipped with Alltek Marine TB-560 beacons shown in Figure~\ref{fig:target-and-drifter}(right) -- were used. These devices transmitted precise GPS positions and velocity vectors at intervals of 10 seconds via VHF radio communication.

%

\subsection{Maritime Search Framework}

\begin{figure}[thpb]
	\centering
	\includegraphics[width=1.0\linewidth]{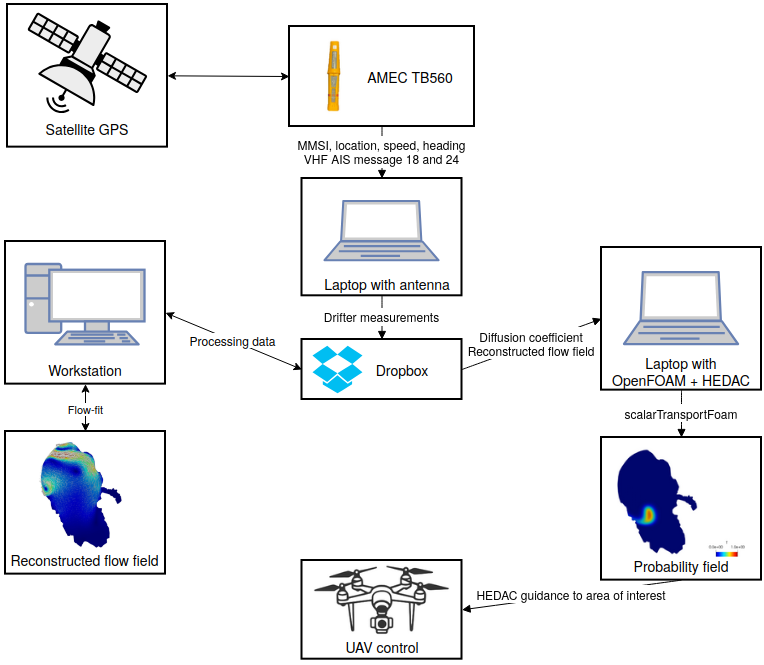}
	\caption{Overview of the search framework integrating drifter data
		processing, flow field reconstruction, and UAV operations to locate objects at sea.}
	\label{fig:search-framework}
\end{figure}

The maritime search framework integrates flow field approximation and UAV operations into a unified system. Drifters equipped with tracking beacons transmit real-time position and velocity data, which are obtained by the land unit and forwarded to a remote workstation for surface flow field approximation. The approximated flow is then forwarded to the UAV Ground Control Station (GCS), where the undetected target probability density field $m$ is advected by the flow field and diffused according to a diffusion coefficient calculated with equation~\eqref{eq:D_c}. Guided by the $m$ field, the UAVs explore the sea surface by capturing images, each of which contributes to the sensing effect applied to the $m$ field within its covered area. After the search is concluded, the captured images are processed using the YOLO detection model to identify target detections. The scheme of the maritime search framework is illustrated in Figure~\ref{fig:search-framework}. 

\subsection{UAV and Sensor Configuration}
The search missions were conducted using either one or two DJI Matrice 210 v2 UAVs, each equipped with a DJI Zenmuse X5S RGB camera capturing 5280 $\times$ 2970~px resolution images. The UAVs operated autonomously at a constant speed of 8 m/s and an altitude of 75 m. The image capture interval was set to three seconds, corresponding to the maximum frequency attainable by the UAV, while still ensuring dense coverage of the search area. Each image captured a sea surface footprint of roughly 95 $\times$ 53.4~m$^2$, sufficient for both accurate detection and efficient area coverage.

The UAVs are controlled at intervals of $\Delta t = 3$ s by a feedback loop that utilizes the potential field obtained from the stationary heat equation~\eqref{eq:hedac_pde}, with parameters $\alpha = 5000$ and $\beta = 0.1$. This feedback loop dynamically refocuses the search effort on high-probability regions as the mission progresses.


\section{Results}

The UAV maritime search field trials consists of three search missions. First two missions utilize the same target and drifter deployment and the same initial probability density. They differ by the delay of the start of the UAV flights and search. For the third mission, the targets and drifters are redeployed to new locations, emulating a different search scenario.

\begin{figure*}[!h]
	\centering
	\includegraphics[width=1.0\linewidth]{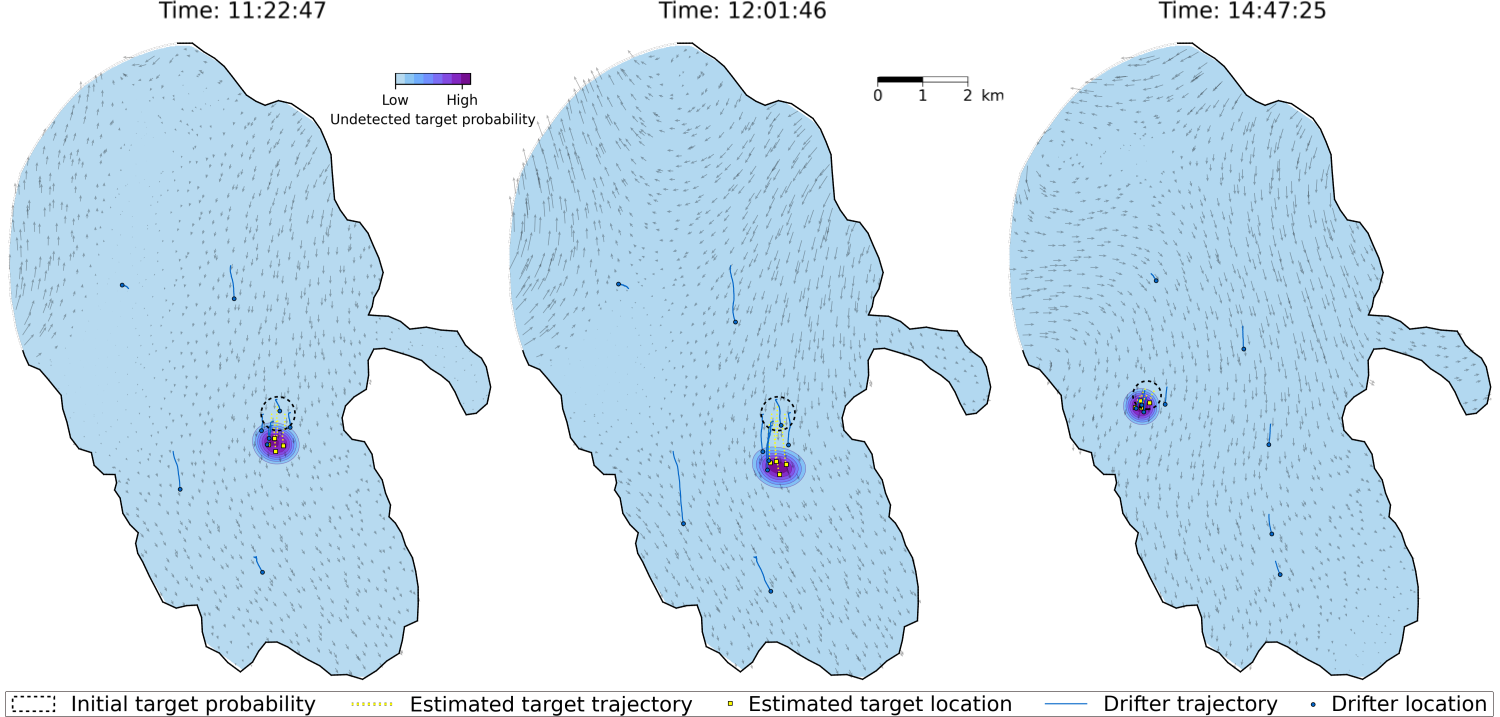}
	\caption{The targets deployment areas (marked with dashed circles) and advected probability distributions (contour plots) at the start of the each search mission. The GPS drifters (blue circle markers) are utilized for flow filed reconstruction (quiver plot). The goal of the missions is to explore the probability distribution in order to find the targets (yellow square markers). Recorded drifter trajectories are marked with blue solid lines while target paths estimated from the reconstructed flow filed are marked as dashed yellow lines.}
	\label{fig:search_t_start}
\end{figure*}

The initial probability density for undetected targets is distributed uniformly within the circular deployment area, then evolved through advection and diffusion under the reconstructed flow for the different delay periods between target deployment and UAV search initiation. Figure~\ref{fig:search_t_start} shows initial deployment locations and flow fields at the start of each mission, as well as monitored trajectories of drifters.

Because the floating targets are not equipped with GPS trackers, their likely positions are estimated using Lagrangian particle advection (Figure~\ref{fig:search_t_start}), as described by the governing equations for motion in a time-varying flow field. Note that the search algorithm only relies on the advected undetected target probability density $m$, while estimated target locations are used in order to demonstrate how difficult is to accurately track a drift of a single particle (target).

\begin{figure*}[!h]
	\centering
	\includegraphics[width=1.0\linewidth]{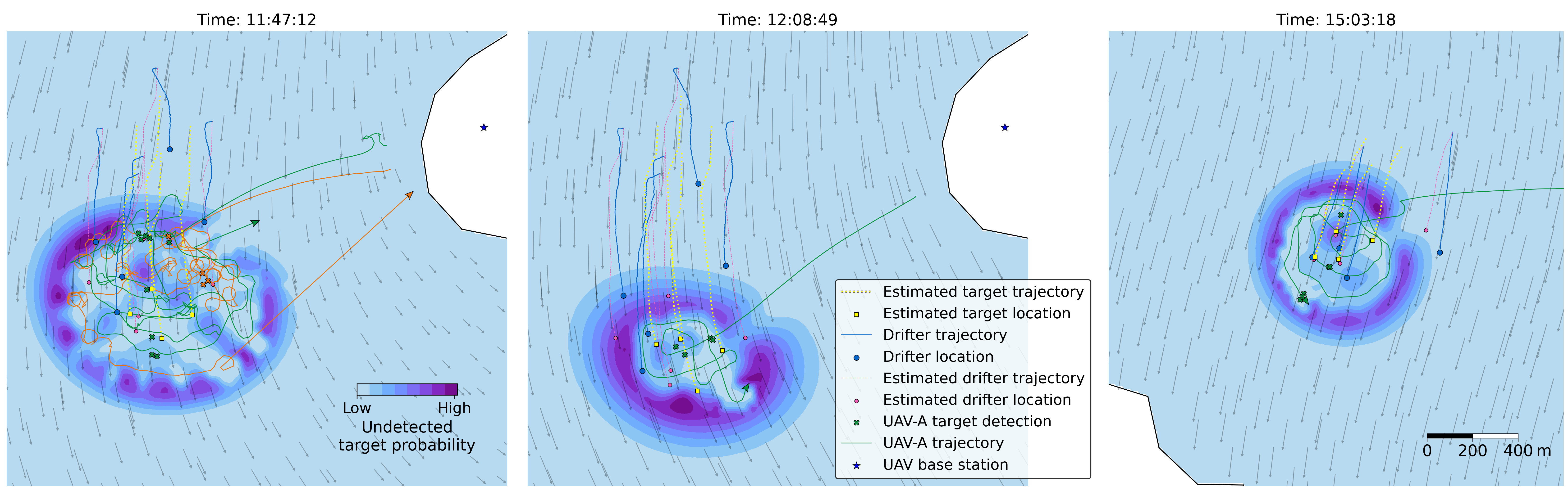}
	\caption{The illustration shows the states of the search at the end of each mission. The real (GPS-measured) trajectories of UAVs and drifters are depicted as solid lines, while their current locations are marked with blue circles and arrows, respectively. Dashed lines indicate the estimated paths of drifters and targets derived from the reconstructed velocity field. Crosses mark the locations of detected targets.}
	\label{fig:search_t_end}
\end{figure*}


The first mission (approx. 11:20-11:45) served as a test for multi-UAV search using proposed framework. The UAVs persistently probed the domain, prioritizing regions where the expected probability of undetected targets remained highest. The meta data of recorded images is sent to the GCS, and this information is forwarded to other subsystems, enabling near-real-time updates of the probability density. Unfortunately, one of the UAVs encountered camera hardware issues, so that only about one in ten images was successfully captured. Nevertheless, the probability field was updated only upon successful image captures, so the mission remained uncompromised.

The second mission (approx. 12:00-12:10) retained the drifter and target configurations from the first mission, with the drifters continuing to drift southward for about an additional hour. The assessed velocity field underwent a noticeable change in flow pattern. The mission was terminated after 10 minutes due to the intermittent signal loss between the UAV and the GCS. 

%

The search area for the third mission (approx. 14:45-15:05) was very distant from the base station, and as a result the mission was short-lived due to battery consumption before single UAV reached the location, resulting with less than 6 minutes of effective search.

At the end of each mission, all recorded images are checked with YOLO detection system, and detected target locations are georeferenced using the UAV telemetry  log and the images' meta data. The results clearly show that targets are successfully registered during the search (see Figure~\ref{fig:search_t_end}). Notably, while most detection locations aligned closely with the estimated target trajectories, some mismatches were observed—reflecting both natural environmental variability and residual errors in the reconstructed flow. Instead of relaying on a particle drift, tracking a target trough its evolving probability distribution allows for a robust search. The adaptive diffusion scheme, anchored in live error estimation from “observer” drifters, effectively compensated for these uncertainties and preserved high correspondence between the modeled and realized distribution of likely target locations. In first mission, due to longest effective search time, multiple detections of all targets are realized. The second mission resulted with only four detection due to camera malfunction, while three of four targets were found in very short search of distant region designated in the third mission.


\section{Conclusion}
This work demonstrates that combining real-time drifter-based surface flow reconstruction, dynamic probability modeling and deep-learning-powered visual detection within an autonomous UAV framework facilitates robust and comprehensive search of floating surface targets under realistic operational conditions. HEDAC UAV motion control proved to be robust, performant, and flawlessly governing the search in presence of environmental dynamics. Field validation demonstrates that the approach can robustly handle uncertainties in models and measurements, while effectively compensating for the inherent limitations of environmental sensing, simulation, and vision-based detection. 

Although this is the initial testing of a very complex system, aside from practical difficulties such as a camera malfunction and occasional loss of radio signal, the results are more than satisfactory.
These results provide a concrete foundation for future developments in autonomous maritime SAR, especially for scenarios where rapid deployment, adaptability, and high detection reliability are required.

\addtolength{\textheight}{-12cm}   



%

\section*{ACKNOWLEDGMENT}

This publication is supported by the Croatian Science Foundation under the project UIP-2020-02-5090 and Univesity of Rijeka under the project uniri-mzi-25-38. \\

\bibliographystyle{ieeetr}  
\bibliography{literature}  

@article{martin2025data,
	title={Data assimilation schemes for ocean forecasting: state of the art},
	author={Martin, Matthew J and Hoteit, Ibrahim and Bertino, Laurent and Moore, Andrew M},
	journal={State of the Planet},
	volume={5},
	pages={1--12},
	year={2025},
	publisher={Copernicus GmbH}
}

@article{lumpkin2013global,
	title={Global ocean surface velocities from drifters: Mean, variance, El Ni{\~n}o--Southern Oscillation response, and seasonal cycle},
	author={Lumpkin, Rick and Johnson, Gregory C},
	journal={Journal of Geophysical Research: Oceans},
	volume={118},
	number={6},
	pages={2992--3006},
	year={2013},
	publisher={Wiley Online Library}
}

@article{lumpkin2017advances,
	title={Advances in the application of surface drifters},
	author={Lumpkin, Rick and {\"O}zg{\"o}kmen, Tamay and Centurioni, Luca},
	journal={Annual review of marine science},
	volume={9},
	number={1},
	pages={59--81},
	year={2017},
	publisher={Annual Reviews}
}

@article{lun2022target,
	title={Target search in dynamic environments with multiple solar-powered UAVs},
	author={Lun, Yuebin and Wang, Honglun and Wu, Jianfa and Liu, Yiheng and Wang, Yanxiang},
	journal={IEEE Transactions on Vehicular Technology},
	volume={71},
	number={9},
	pages={9309--9321},
	year={2022},
	publisher={IEEE}
}

@article{ai2019intelligent,
	title={An intelligent decision algorithm for the generation of maritime search and rescue emergency response plans},
	author={Ai, Bo and Li, Benshuai and Gao, Song and Xu, Jiangling and Shang, Hengshuai},
	journal={Ieee Access},
	volume={7},
	pages={155835--155850},
	year={2019},
	publisher={IEEE}
}

@article{martinez2025maritime,
	title={Maritime search and rescue missions with aerial images: A survey},
	author={Martinez-Esteso, Juan P and Castellanos, Francisco J and Calvo-Zaragoza, Jorge and Gallego, Antonio Javier},
	journal={Computer Science Review},
	volume={57},
	pages={100736},
	year={2025},
	publisher={Elsevier}
}

@article{lanvca2025ergodic,
	title={Ergodic exploration of dynamic distribution},
	author={Lan{\v{c}}a, Luka and Jakac, Karlo and Calinon, Sylvain and Ivi{\'c}, Stefan},
	journal={arXiv preprint arXiv:2503.11235},
	year={2025}
}

@book{gunzburger2012finite,
	title={Finite element methods for viscous incompressible flows: a guide to theory, practice, and algorithms},
	author={Gunzburger, Max D},
	year={2012},
	publisher={Elsevier}
}

@book{lions1996mathematical,
	title={Mathematical topics in fluid mechanics: volume 2: compressible models},
	author={Lions, Pierre-Louis},
	volume={2},
	year={1996},
	publisher={oxford university press}
}

@article{bellomo2015toward,
	title={Toward an integrated HF radar network in the Mediterranean Sea to improve search and rescue and oil spill response: the TOSCA project experience},
	author={Bellomo, Lucio and Griffa, A and Cosoli, S and Falco, Pierpaolo and Gerin, R and Iermano, Ilaria and Kalampokis, Alkiviadis and Kokkini, Z and Lana, A and Magaldi, MG and others},
	journal={Journal of Operational Oceanography},
	volume={8},
	number={2},
	pages={95--107},
	year={2015},
	publisher={Taylor \& Francis}
}

@article{berta2014surface,
	title={Surface transport in the Northeastern Adriatic Sea from FSLE analysis of HF radar measurements},
	author={Berta, Maristella and Ursella, Laura and Nencioli, Francesco and Doglioli, Andrea M and Petrenko, Anne A and Cosoli, Simone},
	journal={Continental Shelf Research},
	volume={77},
	pages={14--23},
	year={2014},
	publisher={Elsevier}
}

@article{JAKAC2025104699,
	title = {Efficient data-driven flow modeling for accurate passive scalar advection in submesoscale domains},
	journal = {Applied Ocean Research},
	volume = {162},
	pages = {104699},
	year = {2025},
	issn = {0141-1187},
	doi = {https://doi.org/10.1016/j.apor.2025.104699},
	url = {https://www.sciencedirect.com/science/article/pii/S0141118725002858},
	author = {Karlo Jakac and Luka Lanča and Ante Sikirica and Stefan Ivić}}

@article{ivic2016ergodicity,
	title={Ergodicity-based cooperative multiagent area coverage via a potential field},
	author={Ivi{\'c}, Stefan and Crnkovi{\'c}, Bojan and Mezi{\'c}, Igor},
	journal={IEEE transactions on cybernetics},
	volume={47},
	number={8},
	pages={1983--1993},
	year={2016},
	publisher={IEEE}
}

\end{document}